\documentclass{article}

\usepackage{arxiv}

\usepackage[utf8]{inputenc} 
\usepackage[T1]{fontenc}    
\usepackage{hyperref}       
\usepackage{url}            
\usepackage{booktabs}       
\usepackage{amsfonts}       
\usepackage{nicefrac}       
\usepackage{microtype}      
\usepackage{lipsum}
\usepackage{graphicx}

\title{Understanding and Creating Art with AI: Review and Outlook}

\author{
	Eva~Cetinic \\
	Rudjer Boskovic Institute\\
	Zagreb\\
	Croatia \\
	\texttt{ecetinic@irb.hr} \\
	\And
	James ~She \\
	\textsuperscript{1} Hong Kong University of Science and Technology\\
	HKUST-NIE Social Media Lab\\
	Hong Kong\\
	\textsuperscript{2} Hamad Bin Khalifa University (HBKU)\\
	Qatar \\
	\texttt{eejames@ust.hk}
}

\begin{document}
\maketitle
\begin{abstract}

Technologies related to artificial intelligence (AI) have a strong impact on the changes of research and creative practices in visual arts. The growing number of research initiatives and creative applications that emerge in the intersection of AI and art, motivates us to examine and discuss the creative and explorative potentials of AI technologies in the context of art. This paper provides an integrated review of two facets of AI and art: 1) AI is used for art analysis and employed on digitized artwork collections; 2) AI is used for creative purposes and generating novel artworks. In the context of AI-related research for art understanding, we present a comprehensive overview of artwork datasets and recent works that address a variety of tasks such as classification, object detection, similarity retrieval, multimodal representations, computational aesthetics, etc. In relation to the role of AI in creating art, we address various practical and theoretical aspects of AI Art and consolidate related works that deal with those topics in detail. Finally, we provide a concise outlook on the future progression and potential impact of AI technologies on our understanding and creation of art. 
\end{abstract}


\section{Introduction}
Recent advances in machine learning have led to an acceleration of interest in research on artificial intelligence (AI). This fostered the exploration of possible applications of AI in various domains and also prompted critical discussions addressing the lack of interpretability, the limits of machine intelligence, potential risks and social challenges. In the exploration of the settings of the ``human versus AI'' relationship, perhaps the most elusive domain of interest is the creation and understanding of art. Many interesting initiatives are emerging at the intersection of AI and art, however comprehension and appreciation of art is still considered to be an exclusively human capability. Rooted in the idea that the existence and meaning of art is indeed inseparable from human-to-human interaction, the motivation behind this paper is to explore how bringing AI in the loop can foster not only advances in the fields of digital art and art history, but also inspire our perspectives on the future of art. 

The variety of activities and research initiatives related to ``AI and Art''  can generally be divided into two categories: 1) AI is used in the process of analyzing existing art; or 2) AI is used in the process of creating new art. In this paper, relevant aspects and contributions of these two categories are discussed, with a particular focus on the relation of AI to visual arts. In recent years, there has been a surge of interest among artists, technologists and researchers in exploring the creative potential of AI technologies. The use of AI in the process of creating visual art was significantly accelerated with the emergence of Generative Adversarial Networks (GAN) \cite{goodfellow2014generative}. The growing number of artists engaging in using AI technologies and the increasing interest from galleries and auction houses in AI Art, fosters discussions about various practical and theoretical aspects of this new movement. On the other side, the increasing online availability of digitized art collections gives new opportunities to analyze the history of art using AI technologies. In particular, the use of Convolutional Neural Networks (CNN) enabled advanced levels of automation in classifying, categorizing and visualizing large collections of artwork image data. Besides building efficient retrieval platforms, smart recommendation systems and advanced tools for exploring digitized art collections, AI technologies can support new knowledge production in the domain of art history by enabling novel ways of analyzing relations between specific artworks or artistic oeuvres. The growing number of artistic productions, research and applications that emerge in the intersection of AI and art, prompts the need to discuss the creative and explorative potentials of AI technologies in the context of our historical, contemporary and future understanding of art. 

\section{Understanding Art with AI} 

\subsection{Art Collections as Data Sources}

\begin{figure}[h]
	\centering
	\includegraphics[width=\linewidth]{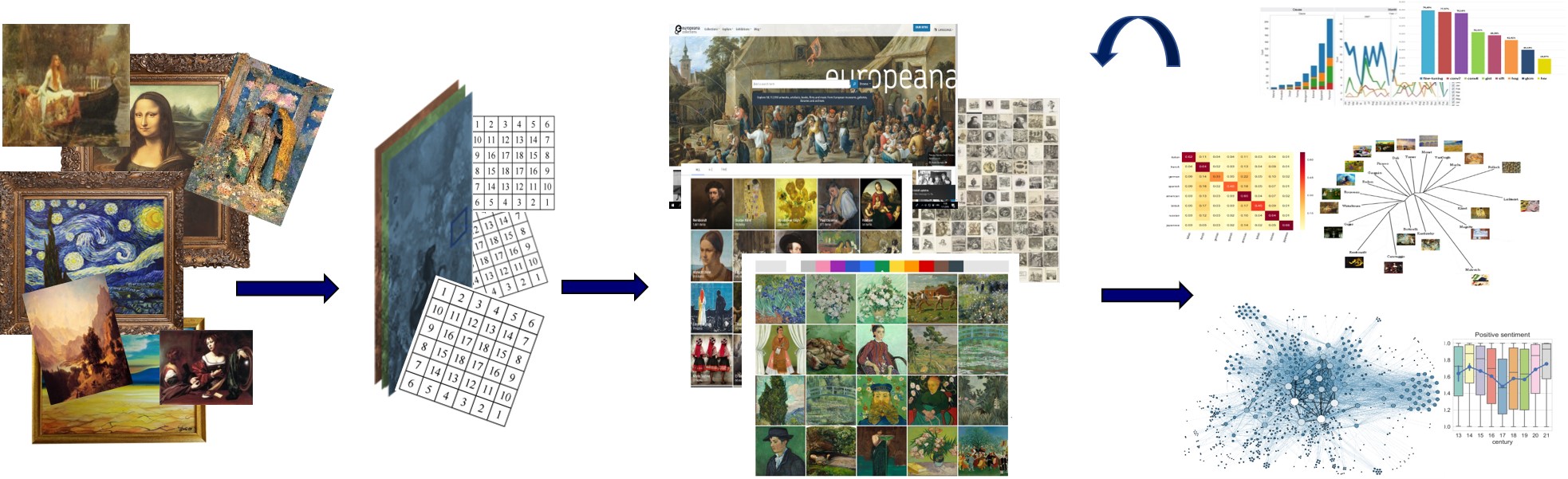}
	\caption{Illustration of the process from digitization to advanced computational analysis.}
	\label{fig1}
	
\end{figure}

Large-scale digitization efforts which took place in the last decades led to a considerable increase of online available art collections. Those art collections enable us to easily explore and enjoy artworks that are located in various museums or art galleries throughout the world. Apart from enabling us to visually inspect various artworks, the availability of large collections of digitized art images triggers new interdisciplinary research perspectives. The physical painting and its digitized counterpart exist in different material modes, yet they encode and communicate the same complex structure of information. Just as the properties of the canvas and the paint often include contextual information that can be of great interest to art historians, the numerical representation of a digitized artwork also includes information of which the potential is not yet fully exploited. The main goal of digitization projects is usually oriented towards building digital repositories to enable easier ways of accessing and exploring collections. Although this is often considered the end goal of many digitization projects, it is important to emphasize that the existence of these collections is only the beginning and necessary prerequisite for applying advanced computational methods and opening new research perspectives. Figure \ref{fig1} illustrates the process from digitization to quantitative analysis, knowledge discovery and visualization using computational methods. Research results obtained from the final phase of this process are often used to enhance the functionalities of repositories and online collections by adding advanced ways of content exploration.

In the context of digitized art analysis, computational methods are generally used either to adopt a distant viewing or a close reading approach \cite{lang2018reflecting}. Close reading implies focusing on specific aspects of one particular work or artistic oeuvre, usually addressing problems such as visual stylometry and computational artist authentication \cite{graham2012statistics, qi2011new, jacobsen2013stylometry, abry2013van}. Most of the studies dedicated to those topics rely on the availability of high-quality digital reproductions of the analyzed artworks and mostly focus on brushstrokes and the texture properties. Distant viewing generally involves analyzing large collections by concentrating on particular features or similarity relations and producing corresponding statistical visualizations. The majority of recent studies concerned with computational analysis of art is exploiting the availability of large digitized collections that include images of various quality, and therefore adopt a distant viewing approach. In the last few years there has been a growing number of research collaborations addressing the application of computer vision and deep learning methods in the domain of digital art history. The most commonly addressed tasks include the problem of automatic classification, object detection, content based and multimodal retrieval, quantitative analysis of different features and concepts, computational aesthetics, etc. 

The availability of large-scale and well-annotated datasets is a necessary requirement for adopting deep learning models for various tasks. In recent years, many museums and galleries published digital online versions of their collections. Since it is difficult to list all the existing online art collections, this paper focuses only on those collections and datasets that have been most commonly used for computer vision and deep learning research in the last few years.  
Table \ref{tab1} contains a list of the most well-known collections of primarily Western art that were often used as sources for creating different task-specific datasets. Also, it includes well-known and novel datasets that have been specifically developed for a particular research task.

\begin{table*}[h]
	\centering
	\caption{List of artwork datasets associated with their corresponding number of images and tasks for which they are designed or most commonly used. }
	\begin{tabular}{p{0.23\textwidth}p{0.18\textwidth}p{0.1\textwidth}p{0.35\textwidth}}
		\toprule
		Dataset                 & Source                                 & Number of images & Task                                                                                   \\ \midrule
		Painting-91             & \cite{khan2014painting}                & 4k               & artist   classification                                                                \\
		Pandora                 & \cite{florea2016pandora}               & 8k               & style   classification                                                                 \\
        Web Gallery of Art      & \url{www.wga.hu}                       & 40k             & artist, style, period classification \\
        WikiArt                 & \url{www.wikiart.org}                  & 85k           & artist, style, genre   classification \\
		Rijksmuseum Challenge   & \cite{mensink2014rijksmuseum}          & 112k             & artist, material, type classification                                                \\
		Art500k                 & \cite{mao2017deepart}                  & 550k             & artist, genre, style, event, historical figure retrieval                            \\
		OmniArt                 & \cite{strezoski2018omniart}            & 2M               & artist, style, period, type, iconography, color classification / object detection    \\
		ArtDL                   & \cite{milani2020data}                  & 42k              & iconographic  classification                                                          \\
		PRINTART                & \cite{carneiro2012artistic}            & 1k               & object/ pose retrieval                                                                 \\
		Paintings               & \cite{crowley2014state}                & 8.6k             & object retrieval                                                                     \\
		Face Paintings          & \cite{crowley2015face}                 & 14 k             & face retrieval                                                                         \\
		Iconart                 & \cite{gonthier2018weakly}              & 6k               & iconographic object   detection                                                        \\
		VisualLink              & \cite{seguin2016visual}                & 38.5k            & visual link   retrieval                                                                \\
		Brueghel dataset        & \cite{shen2019discovering}             & 1.6k             & visual link   retrieval                                                                \\
		IconClass AI Test Set & \cite{posthumus2020brill} & 87k              & iconographic   classification / multimodal tasks                                       \\
		SemArt                  & \cite{garcia2018read}                  & 21k              & multimodal  retrieval                                                                 \\
		Artpedia                & \cite{stefanini2019artpedia}           & 3k               & multimodal  retrieval                                                                 \\
		BibleVSA                & \cite{baraldi2018aligning}             & 2.3k             & multimodal  retrieval                                                                 \\
		AQUA                    & \cite{garcia2020dataset}               & 21k              & visual question answering                                                            \\
		ArtEmis                 & \cite{achlioptas2021artemis}           & 81K              & multimodal sentiment analysis                                                        \\
		WikiArt Emotions        & \cite{mohammad2018wikiart}             & 4.1k             & sentiment analysis/ emotion classification                                           \\
		MART                    & \cite{yanulevskaya2012eye}             & 500              & sentiment analysis                                                                     \\
		JenAesthetic            & \cite{amirshahi2014jenaesthetics}      & 1.6k             & aesthetics quality   assessment                                                        \\ \bottomrule
	\end{tabular}
	\caption{List of artwork datasets associated with their corresponding number of images and tasks for which they are designed or most commonly used. }
	\label{tab1}
\end{table*}

The datasets in Table ~\ref{tab1} are linked to tasks for which they have been designed or most commonly used. The majority of online art collections include general annotations related to the whole image and are often used for classification or retrieval tasks. Those annotations are mostly provided by art experts and contain information about the artist, style, genre, technique, period, etc. Some specific tasks such as object detection require more detailed annotations of specific image regions. Recently several datasets of images associated with textual descriptions have emerged in order to perform different multimodal tasks. The datasets used for sentiment analysis and aesthetic quality assessment usually contain annotations that were collected from multiple annotators through specific surveys or crowdsourcing platforms.

\subsection{Automated Classification of Artworks}

Automated classification of artworks based on categories such as artist, style or genre has been one of the central challenges of computational art analysis over the last decade. Most of the earlier studies addressed the problem of automatic artist \cite{Keren02, Cetinic2013artist}, style \cite{Shamir2010, Shamir2012} and genre classification \cite{Agarwal2015} by extracting various hand-crafted image features and employing different machine learning algorithms using those features. In achieving better classification accuracy, momentous progress has been made with the adoption of convolutional neural networks (CNNs). In the beginning, CNNs were first employed as feature extractors. Karayev et al. were the first to utilize layer activations of a CNN trained on ImageNet \cite{deng2009imagenet}, a large hand-labelled object dataset, as features for artistic style classification \cite{Karayev2014}. In their work they showed that features extracted from a network trained for a completely different task (object recognition on a natural image dataset) outperformed all other low-level image features on the task of style classification. The dominance of CNN-based features, particularly in combination with other hand-crafted features, was confirmed for artist \cite{David2016}, style \cite{Bar2014} and genre classification \cite{Cetinic2016genre}. Apart from using pre-trained CNNs as just feature extractors, Girshick et al. showed that further improvement of performance for a variety of visual recognition tasks can be achieved by fine-tuning a pre-trained network on the new target dataset \cite{Girshick2014}. Both CNN feature extraction and fine-tuning represent forms of transfer learning, where knowledge that a model learned on one task is being exploited for a new task. Transfer learning approaches, particularly fine-tuning, have proven to give state-of-the-art results for different artistic datasets and various classification tasks \cite{Noord2017,Lecoutre2017, sandoval2019two, yang2020classification, bianco2019multitask, menis2020deep}. 
In order to better understand the transferability of pre-trained models, Cetinic et al. \cite{cetinic2018fine} explore how different fine-tuning strategies and domain-specific model initializations influence the classification performance of various art classification tasks and datasets when using the same CNN architecture. Sabatelli et al. \cite{sabatelli2018deep} analyze the transferability and fine-tuning impact of different CNN architectures and conclude, similarly as Gothier et. al \cite{gonthier2020analysis} who also perform a transfer learning analysis, that models fine-tuned on art datasets outperform ImageNet pre-trained models when applied on different tasks and dataset from the art domain. A comprehensive overview of the current work related to classification of artworks, as well as an approach towards classifying artworks based on iconographic elements is presented in \cite{milani2020data}.

\subsection{Object Detection and Similarity Retrieval}

Besides classification, the use of deep neural networks showed promising results in exploring the content of artworks and automatically recognizing objects, faces or other specific motifs in paintings. As one of the pioneering works in this area was, Crowley et al. \cite{crowley2014search} showed that object classifiers trained using CNN features from natural images can retrieve paintings containing these objects with great success. Later studies addressed the problem of not only retrieving paintings that depict a specific object, but also determining the position of the object in the image \cite{crowley2016art, gonthier2018weakly}, as well as detecting content to discover co-occurring patterns in collections \cite{kim2019computational}. One aspect of content that gained particular attention is the depiction of human faces. Interesting work has been done on the topic of people and face detection in paintings \cite{wechsler2019modern, mzoughi2018face, mermet2020face, westlake2016detecting, mermet2020face}, as well as analysis and classification of the detected faces based on gender and other features \cite{strezoski2018omniart}. Apart from faces, effort has been made to recognize other content-related elements of artworks, such as detecting the pose of characters in paintings \cite{jenicek2019linking, madhu2020understanding, bell2019ikonographie}, recognizing specific characters \cite{madhu2019recognizing} or detecting materials depicted in paintings \cite{Lin2020insights}. 

One of the main practical goals for employing computational methods for automated content and style recognition in art images is to build smart retrieval systems that can help organize and analyze large collections of artworks in an efficient way. Many of the existing retrieval systems rely on retrieving images based on their corresponding metadata and textual descriptions. However, with convolutional neural networks, significant progress has been made towards  obtaining relevant results with image-based enquiries. The notion of ``visual similarity'' is often considered a key factor in various retrieval systems. However, in the context of art, ``similarity'' is a complex term that can include different aspects of content matching (depiction of the same objects or similar iconographic representations) or correspondences that are more related to style such as brushstrokes, color, composition, etc. The application of CNNs has shown promising results in retrieving visually linked images in painting collections \cite{seguin2016visual, shen2019discovering, castellano2020towards, castellano2020visual}. To tackle the complexity of similarity in art, Mao et al. \cite{mao2017deepart} proposed the DeepArt retrieval system that encodes joint representations that can simultaneously capture content and style features. A more detailed discussion of the problem of similarity and applications of computer vision methods for the organization and study of art image collections is given in \cite{lang2018attesting}. 

\subsection{Multimodal Tasks}

Besides exploring only visual similarities, recently an increased number of studies focused on analyzing both visual and textual modalities of artwork collections. Efforts to map images and their textual descriptions in a joint semantic space have mostly been made in order to create multimodal retrieval systems. In particular, Garcia and Vogiatzis \cite{garcia2018read} introduced the SemArt dataset, a collection of fine-art images associated with textual descriptions, and applied different methods of multi-modal transformation with the goal to map the images and their descriptions in a joint semantic space. Baraldi et al. \cite{baraldi2018aligning} introduced a new dataset named BibleVSA, a collection of miniature illustrations and commentary text pairs, to explore supervised and semi-supervised approaches of learning cross-references between textual and visual information in documents. Stefanini et al. \cite{stefanini2019artpedia} presented the Artpedia dataset where images are annotated with both visual and contextual descriptions and introduced a retrieval model that maps images and sentences in a joint embedding space and discriminates between contextual and visual sentences of the same image.

Besides retrieval, another topic of interest in the domain of multimodal tasks is visual question answering (VQA). VQA refers to the problem where given an image and textual question, the task is to provide an accurate textual answer. Bongini et al. \cite{bongini2020visual} annotated a subset of the Artpedia dataset with visual and contextual question-answer pairs. They introduced a question classifier that discriminates between visual and contextual questions and a model capable of answering both types of questions. Garcia et al. \cite{garcia2020dataset} presented a novel dataset AQUA, which consists of automatically generated visual- and knowledge-based QA pairs, and introduced a two-branch model where the visual and knowledge questions are managed independently. Apart from VAQ, a few recent works addressed the task of image captioning where the goal is to automatically generate accurate textual descriptions of images. Sheng and Moens \cite{sheng2019generating} introduce image captioning datasets referring to ancient Egyptian and Chinese art and employ an encoder-decoder framework for image captioning where the encoder is a CNN and the decoder is a long short-term memory (LSTM) network. In the context of art history, it would be preferable to generate not only simple descriptions of the image content, but more advanced descriptions that articulate the theme and symbolic associations between objects. A step in this direction was presented in \cite{cetinic2021iconographic} where the image-text pairs from the Iconclass AI Test Set dataset were used to fine-tune a transformer-based vision-language pre-trained model for the task of iconographic image captioning. 

\subsection{Knowledge Discovery in Art History}

Apart from creating practically useful retrieval systems, another important contribution of computational analysis of art is the opportunity to adopt a quantitative approach in studying theoretical concepts relevant for art history. Elgammal et al. \cite{elgammal2018shape} showed that internal representation of convolutional neural networks trained for style classification encode discriminative features that are correlated with art historical concepts of style transformation. Later studies expanded this research direction by analyzing the semantic interpretation of deep neural network  features in relation to different visual attributes \cite{kim2018finding} and their employment for quantifying and predicting values of art historically relevant stylistic properties \cite{cetinic2020learning}. One of the most interesting aspects of adopting a quantitative approach in analyzing large artistic datasets is the possibility to define high-level features that correspond to abstract notions of understanding art. For example, Deng et al. \cite{deng2020exploring} introduce the notion of ``representativity'', that indicates the extent to which a painting is considered typical in the context of an artist’s oeuvre, using deep neural networks to extract style-enhanced image representations. It is necessary to emphasize that there is a wide range of studies related to quantitative analysis of art that do not adopt deep learning based methods, but employ image processing and statistical methods from information theory \cite{kim2014large, sigaki2018history, lee2020dissecting}. 

Particularly important for reaching the goal of advanced computational analysis of art is a stronger collaboration between different disciplines, especially computer science and art history. In the context of computer science, the topics related to art history are often addressed on a superficial level and art collections are treated as just another source of image datasets. On the other hand, art historians tend to refrain from embracing computational methods in their research due to practical difficulties, cross-domain knowledge gaps or animosity towards the increasing trend of quantification in humanities research. However, in the recent years a growing number of art history research projects are adopting computational methods and digital art history is becoming an established field \cite{klinke2020digital}. Also, research criteria are becoming more rigorous and computational methods are not being used only because it is fashionable, but because  they can provide truly novel methodological extensions. Interdisciplinary research is not only necessary to better understand how computational methods, particularly deep learning techniques, can support research in digital art history, but also how research questions that are relevant in art history can foster new challenges for artificial intelligence. Because of its manifold nature, collections of artworks represent a prolific data source for formulating various complex tasks related to computational image understanding, multimodal analysis, affective computing and computational creativity.  

\subsection{Aesthetics and Perception}

Perhaps the most intriguing, but at the same time the most ambiguous, topic in the context of computational analysis of art is the one related to perception. Different aspects of visual perception have been studied by psychologists for a long time and have in the recent years become an rising subject of interest within the computer vision and deep learning community. In particular, computational aesthetics is a growing field preoccupied with developing computational methods that can predict aesthetic judgments in a similar manner as humans. Developing quantitative methods for analyzing subjective aspects of perception is particularly challenging in the context of art images. One of the major challenges in studying perceptual characteristics of images is the development of large-scale datasets annotated with evaluation scores obtained using experimental surveys. Amirshahi et al. introduced the JenAesthetics dataset \cite{amirshahi2014jenaesthetics}, a datasets of artworks images labelled with subjective scores of aesthetic evaluation. Several studies have addressed the topic of computational aesthetics in art by analyzing various statistical properties of paintings \cite{hayn2017subjective, sargentis2020aesthetical, khalili2021information}. Zhao et al. \cite{zhao2020representation} propose a method for CNN-based image composition features and how they can be used for aesthetic prediction in natural and art image datasets. In order to include other aspects of perception, Cetinic et al. \cite{cetinic2019deep} used deep learning based quantitative methods to extract features not only related  to aesthetic evaluation, but also to the sentiment and the memorability of fine art images. Besides aesthetic evaluation, sentiment analysis is the most commonly addressed task in this domain. Mohammad and Kiritchenko \cite{mohammad2018wikiart} introduced WikiArt Emotions, a dataset of paintings that has annotations for various emotions evoked in the observer. Alameda-Pineda et al. \cite{alameda2016recognizing} introduced an approach to automatically recognize the emotion elicited by abstract paintings using the MART dataset \cite{yanulevskaya2012eye}, a collection of 500 abstract paintings labelled as evoking positive or negative sentiment. Most recently, Achlioptas et al. \cite{achlioptas2021artemis} introduced ArtEmis, a large-scale dataset of emotional reactions to visual artwork joined with explanations of these emotions in language, and developed machine learning models for dominant emotion prediction from images or text. Developing computational approaches for quantifying and predicting values of concepts such as aesthetics or sentiment is especially difficult for art images. The importance and originality, and therefore also our perception, of a particular artwork do not only emerge from its visual properties, but greatly depend on the art historical context. For that reason, it is obvious that current approaches are limited because they only take into account visual image features. This also indicates that future research has to aim towards a more holistic approach if we ought to build systems that can achieve a human-like understanding of art.

\section{Creating AI Art}

\subsection{Technological Milestones}

\begin{figure}[h]
	\centering
	\includegraphics[width=\linewidth]{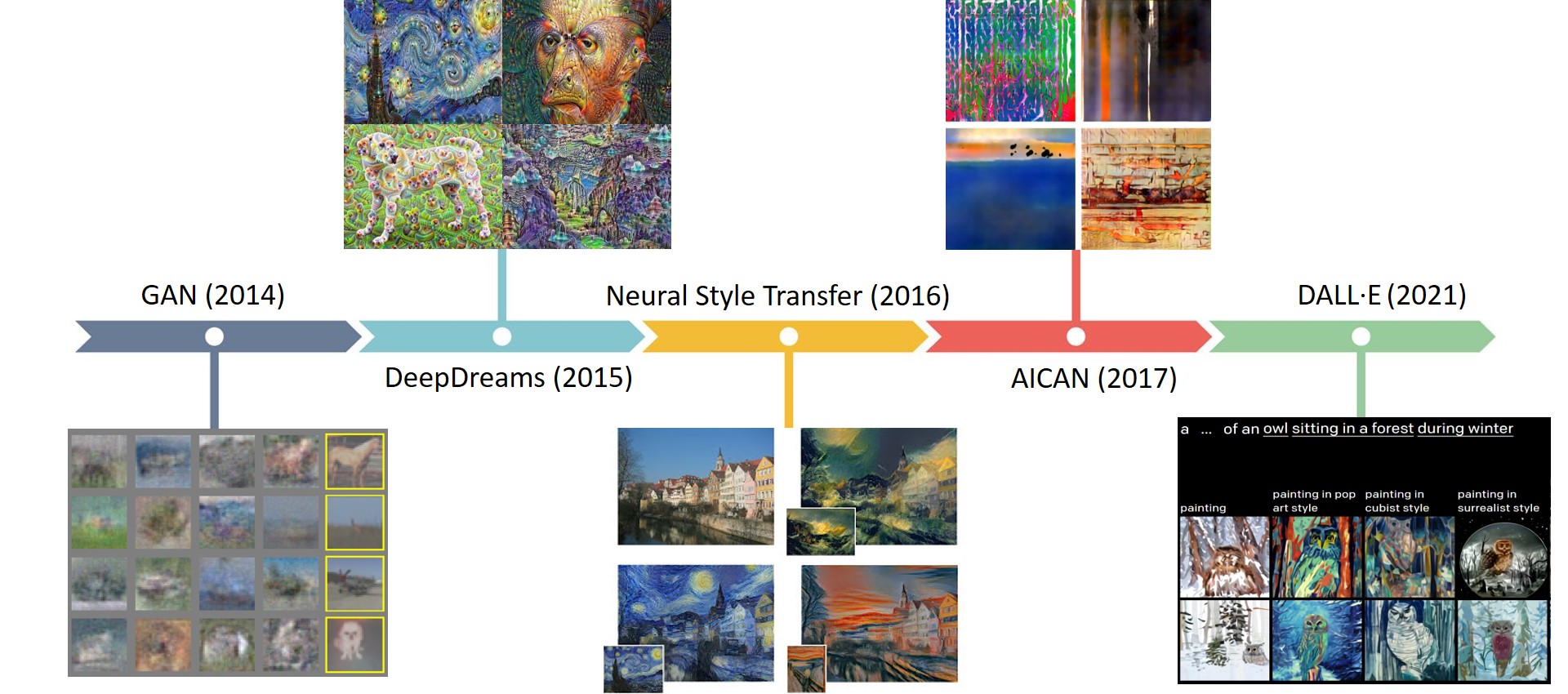}
	\caption{Ilustration of the most important technological milestones that led to the current AI Art production.}
	\label{fig2}
	
\end{figure}

Several important technological advances achieved in the last few years supported the rising interest in AI Art. In the context of computer graphics and computer vision research, over the last decades many rendering and texture synthesis algorithms have been developed. Those algorithms were designed to modify images in various ways, including the application of an ``artistic style'' to the input image, e.g. painterly or sketched style \cite{hertzmann1998painterly,hertzmann2001image,efros2001image,gooch2001non}. However, the use of deep neural networks for the purpose of stylizing photos and creating new images began rather recently and accelerated in the last five years. Figure \ref{fig2} summarizes some of the most important technological milestones that influenced AI Art production. One of the first methods that gained significant attention was DeepDreams introduced by Mordvintsev et al. \cite{mordvintsev2016inceptionism} in 2015. This method was initially designed to advance the interpretability of deep convolutional neural networks by visualizing patterns that maximize the activation of neurons. Because it produced a rather psychedelic and hallucinatory stylistic effect, the method later became a popular new form of digital art production.   

One of the most iconic AI inventions that triggered the rapid use and development of AI technologies for art was Neural Style Transfer (NST). This method was introduced in the highly influential work of Gatys et al. \cite{gatys2016image} that demonstrated the successful use of CNNs in creating stylized images by separating and combining the image ``content'' and ``style''. This breakthrough was followed by many new research contributions and applications, a comprehensive overview of the existing NST techniques and their various applications is given in \cite{jing2019neural}. Terminology of computer graphics and computer vision suggests that ``content'' and ``style'' are understood in a rather straightforward and simple manner. ``Content'' is associated with recognizable objects and figures that are depicted in an image, while ``style'' implies an aesthetically pleasing or interesting visual deviation from the photorealistic depiction of content. However, in the context of art history, style is not only associated with mere visual characteristics of lines and brushstrokes, but is often considered a subtle and contextually dependent concept. Furthermore, stylized images produced using NST methods most commonly represent an obvious combination of existing image inputs and not an original and unique artistic creation. NST methods surely represent a very interesting technological contribution in the domain of automated image manipulation. It is therefore understandable that many applications relying on those methods emerged in order to enable end-users an easy and entertaining framework for photo manipulation. Although NST do have the potential to be used in a creative way for digital art production, it is also necessary to keep a critical stance towards the trend of proclaiming everything with a painterly overlay as art. Due to the limitless combinations, it is indeed technically challenging and time-consuming to locate and pair up two matching content and style images that could produce an aesthetic, meaningful and memorable output image as a novel artwork. 

The perhaps most relevant technological innovation that contributed significantly to the current rise of the AI Art movement are Generative Adversarial Networks (GANs). Introduced by Goodfellow et al. \cite{goodfellow2014generative}, GANs represent a turning point in the attempt to use machines for generating novel visual content. The key mechanism of a GAN is to train two ``competing'' models that are usually implemented as neural networks: a generator and a discriminator. The goal of the generator is to capture the distribution of true examples of the input sample and generate realistic images, while the discriminator is trained to classify generated images as fake and the real images from the original sample as real. Designed as a minimax optimization problem, the optimization process ends at a saddle point that is considered a minimum in relation to the generator and a maximum to the discriminator. The implementation of this framework showed impressive results in generating convincing fake variations of realistic images for various types of image content. GAN soon became one of the most important research area in artificial intelligence and many advanced and domain-specific variations of original architecture emerged, e.g. CycleGAN \cite{zhu2017unpaired}, StyleGAN \cite{karras2019style} or BigGAN \cite{brock2018large}. 

To take the GAN technology one step further in its capacity to generate content in a creative way, Elgammal et al. \cite{elgammal2017can} have introduced AICAN - artificial intelligence creative adversarial network. In their article they argue that if a GAN model is trained on images of paintings it will just learn how to generate images that look like already existing art, and in a similar manner as the Neural Style Transfer method, this will not produce anything truly artistic or novel. In their paper they propose modifications to the optimization criterion to enable the network to generate creative art by maximizing deviation from established styles while staying within the art distribution. Through a series of exhibitions and experiments, authors of the AICAN system showed that people were very often unable to tell the difference between AICAN-generated images and artworks produced by a human artist \cite{elgammal2019ai}. Besides the AICAN initiative, in order to generate their digital artwork, many other developers and artists employed GANs with various modifications and specific training settings and it has become the most widely used technology in the current AI Art scene. 

In the meantime, within the AI research community there has been a surge of interest in transformer-based architectures \cite{vaswani2017attention} and their successful application to various tasks in different areas, with a particular emphasis on text and multimodal applications. In January 2021 OpenAI presented a very advanced neural network called DALL·E that creates images from text captions for a wide range of concepts expressible in natural language \cite{openaidelle}. While there have been many attempts to create text-to-image synthesis systems \cite{reed2016generative, xu2018attngan}, the newly presented DALL·E results seem very promising and have recently gained a lot of attention. Although this specific model is currently not openly available for use, we assume that advanced text-to-image synthesis models such as this one will represent an important trend in the future of AI Art. 

\subsection{The Contemporary AI Art Scene}

The universal rule of triggering attention by controversy has once again proven effective in the case of AI Art. Since October 2018, when the AI artwork ``Portrait of Edmond Belamy'' produced by the Obvious collective was sold at an auction by Christie’s for US\$432,500 \cite{aiartarticle1} there has been an increasing interest for AI Art, but also a growing need to discuss key aspects of this new movement in the contemporary art scene. The case of the ``Portrait of Edmond Belamy'' particularly provoked the discussion about authorship and ethics. However, other important questions started gaining attention from art historians and artists, as well as AI scientists and developers, such as questions related to novelty, originality and autonomy in AI Art.  

\begin{figure}[h]
	\centering
	\includegraphics[width=\linewidth]{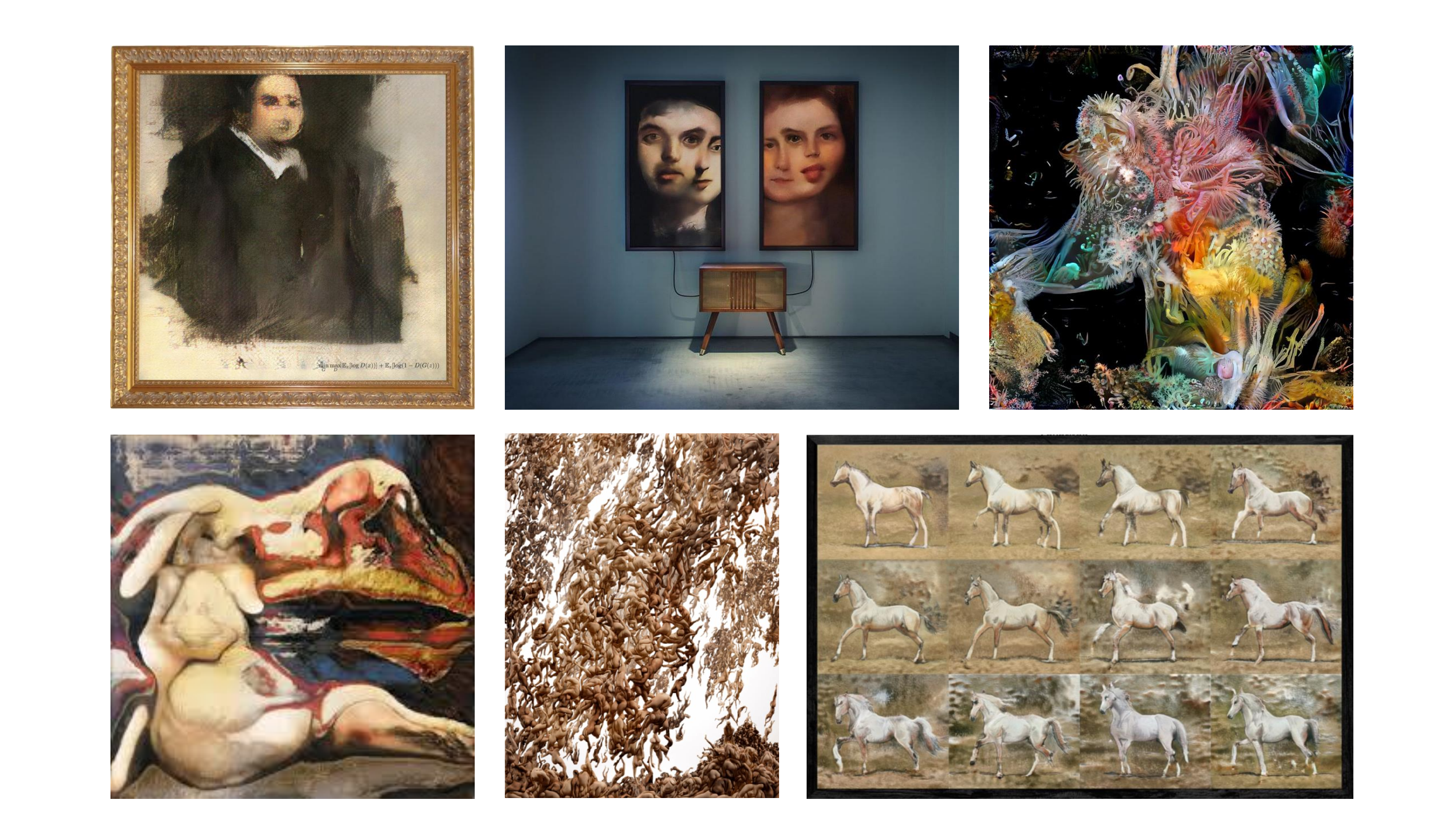}
	\caption{Examples of AI Art: 1) Obvious, Portrait of Belami 2) Mario Klingemann, Memories of Passersby I 3) Sofia Crespo, Neural Zoo 4) Robbie Barrat, Nudes 5) Scot Eaton,  Humanity (Fall of the Damned) 6) James She, Keep Running}
	\label{fig3}
	
\end{figure}

Although the case of the ``Portrait of Edmond Belamy'' became very popular for various reasons, within the AI Art community many agree that there are other artists whose work can be considered a more representative example of AI Art: ``Many artists working in the field point out that Obvious, the collective behind the AI that made the pricey work, was handsomely rewarded for an idea that was neither very original nor very interesting'' \cite{aiartarticle2}. However, since the boost of AI Art popularity triggered by Christie's auction, the number of artists involved in making AI Art is rising worldwide. The fact that AI Art gained momentum in the last two years is reflected in the growing number of online platforms featuring AI Art such as AIArtists.org \footnote{\url {https://aiartists.org/}}, as well as exhibitions, conferences, competitions and discussion panels dedicated to AI Art. Figure \ref{fig3} shows several examples of contemporary AI artworks. 

Although various artworks are labelled as ``AI Art'', it is not often obvious what exact AI technologies are used in the production of specific artworks, as many artists do not reveal all details of their creative process. However, because AI Art production is rapidly increasing and gaining attention, it is necessary to understand and discuss all the aspects that play a role in judging the quality of a particular work. A certain level of technical knowledge and skillfulness is currently required in order to participate in the practice. However, applications of AI technologies are rapidly advancing towards more user-friendly and easily operated frameworks. It is therefore difficult to estimate if the value of a particular AI artwork should depend on the technological complexity and innovation involved in its production, or only on the final visual manifestation and contextual novelty.

\subsection{Novelty of AI Art}

The growing trend of using AI technology in art production, triggers discussions about the fundamental questions regarding the artistic nature of these works and their place in the history of visual arts. With the quest for understanding the dynamics of AI Art, it is necessary to address the issue of novelty of this type of art in the context of art history. Is the 21st century just delivering now technologically possible solutions to ideas conceived way back in the 20th century? The anticipation of the forms that are now coming to life using advanced technology has been articulated a long time ago. To understand the novelty of AI Art in its current form, it is important to acknowledge that the application of computers to arts started with the earliest days of computing. But even before the use of computers, ideas related to uncertainty and the simulation of chance have already been artistically articulated, e.g. the ``action paintings'' by Jackson Pollock or the ``chance collages'' by Jean Arp. In order to gain a better understanding of the historical context of AI Art, we refer the reader to an overview of chance-assisted creativity \cite{dorin2013chance} and stochastic process in art \cite{risset1982stochastic}. Furthermore, ``generative art'', understood as a concept describing the employment of a system with some degree of autonomy as a relevant component in the production of art, has been extensively theoretically and practically explored in the last several decades \cite{galanter2003generative, boden2009generative, mccormack2014ten, dorin2012framework}. Initiatives such as the ``The Painting Fool'' project, that started in 2006, can be considered predecessors of the current AI Art movement, in the sense that it not only pursued the technological challenges but also tackled the sociological aspect of its goal to build a software that  ``will one day be taken seriously as a creative artist in its own right'' \cite{colton2012painting}  

Because of the rising attention that AI Art gained in the last few years and the overall ``hype'' related to everything associated with the acronym ``AI'', many scholars discussing AI Art found it necessary to emphasise its historical context. Elgammal \cite{elgammal2019ai} reminds us that Harold Cohen created one of the first programs for computer-generated art in 1973. The program is called AARON and was used to produce drawings that followed a predefined set of rules. Todorov \cite{todorov2019game} argues that processes resembling what AI currently does when generating art have already been expressed without the use of computers. He mentions the example of the book by Raymond Queneau, Cent Mille Milliards de Poèmes published in 1961, which was structured and printed so that the reader could create $10^{14}$ different combinations of poems.  A very thorough and comprehensive discussion of AI Art in the context of visual art history is presented by Aaron Hertzmann in his essay ``Can computers create art?'' \cite{hertzmann2018can}. In his article, Hertzmann draws parallels between AI Art and the invention of photography, as well as explores the evolution of collaboration between art and technology in filmmaking, 3D computer animation and procedural artwork. 

Having in mind the historical background, it is understandable to question the originality and novelty of the fundamental principles of AI Art. Nevertheless, the specific technological innovations that emerged in the last few years, did open possibilities for exploring those principles on a different scale. Most of the current AI Art works can be understood as results of sampling the ``latent space''. Perhaps the most novel aspect of AI Art is this possibility to venture into that abstract multi-dimensional space of encoded image representations. From the artist’s perspective, the latent space is neither a space of reality nor imagination, but a realm of endless suggestions that emerge from the multi-dimensional interplay of the known and unknown. How one orchestrates the design of this space and what one finds in it, eventually becomes the major task and distinctive ``signature'' of the artist. In this context, it is important to understand the role of the human in this collaborative process with the machine. 

\subsection{Machine Autonomy and the Role of the Artist}

There has been an ongoing debate about the human-computer relationship in the process of creating AI Art. One of the fundamental questions is related to the level of autonomy the computer has in making decisions that can be considered essential for the creative process. Are computational technologies still regarded as mere tools or do they exhibit properties of independent ``behaviour''? And is the framing of a specific narrative grounded in the reality of the procedure or in the underlying marketing reasons?

J. McCormack et al. \cite{mccormack2019autonomy} discuss the question of autonomy of GANs in relation to the comprehensive study of autonomy in computer art presented in 2010 by Boden \cite{boden2010creativity}. They argue that GANs have a very limited capacity for autonomy because they synthesize images that mimic the latent space of the training data, but do not have any role in choosing the input datasets nor the statistical model that represents the latent space. In this sense, the authors suggest that ``GANism is more a process of mimicry than intelligence'' and that its capacity for autonomy is not significantly higher than the one of prior generative systems. Mazzone and Elgammal \cite{mazzone2019art} agree that many of the recent GAN produced artworks use AI as a tool, while the creative process is primarily dependent on the artist’s pre- and post-curatorial actions. However, they describe the AICAN system as an ``almost autonomous artist'' and state that unlike in the case of previous generative art, the process behind AICAN is inherently creative. One of their main arguments for this claim is the fact that they did not perform any curation on the input dataset, but used 80K images of various genres and styles from the Western canon in order to simulate the process of how an artist absorbs art history. Also, the optimization during training of the network was performed so that the final output represents an optimal point between mimicking and deviating from existing styles. Hertzmann on the other site, indicates ``that AI algorithms are not autonomous creators and will not be in the foreseeable future. They are still just tools, ready for artists to explore and exploit.'' \cite{hertzmann2018can}. He claims that systems such as AICAN cannot be compared to human artists because they do not grow or evolve over time. He argues that although it would be rather easy to integrate certain mechanisms of change within current systems, achieving a truly meaningful evolution while at the same time producing art that is relevant to the human audience, is still a very distant scenario. In his recent work ``Computers Do Not Make Art, People Do'' \cite{hertzmann2020computers}, Hertzmann emphasizes his position by stating that describing AI systems as autonomous artists is irresponsible because it can mislead people into thinking that those systems have human-like attributes such as intelligence and emotions. 

In an age of information inflation and hyper-production of art, gaining attention has become one of the most important principles of success. Considering the current hype around AI, it is understandable that framing the story behind a particular artwork as something being made autonomously by an AI system, seems to trigger more interest than yet another human-made work. Notaro \cite{notaro2020state} argues that the narrative of the ``autonomous AI artist'' is driven by marketing reasons and that exhibitions of works made by AICAN, similarly as in the case of Christie’s Belamy auction, exploit the idea of autonomy for the sake of publicity. Epstein et al. \cite{epstein2020gets} also point out that the employment of anthropomorphic language in the case of the Christie’s Belamy auction significantly increased the public interest in the work. The authors provide a detailed analysis of the media coverage and its role in creating a discourse that emphasized the autonomy of the algorithm. Nevertheless, as AI technologies are becoming more and more sophisticated, the distinction between using a AI system as a tool or as a creator of content is becoming more vague. Having in mind that our overall understanding and interpretability of AI systems is limited, initiatives such as Explainable Computational Creativity (XCC), as a subfield of Explainable AI (XAI) \cite{llano2020explainable}, are becoming very relevant as future research directions. Also, the complexity of this topic exceeds the boundaries of one particular research area and is starting to gain more attention from scholars from various disciplines. Coeckelbergh \cite{coeckelbergh2017can} presented a conceptual framework for philosophical thinking about machine art by analyzing from various perspectives the question if machines can create art. While providing a very detailed analysis of the question of autonomy, Daniele and Song \cite{daniele2019ai+} also indicate the importance of interdisciplinary dialogue and point out that discussions should not only be limited to the question if machines can create ``real" art, but also consider the social and cultural implications of interacting with art that was autonomously generated by non-human creative systems. Therefore, besides understanding the technological, artistic or philosophical aspects of AI Art,  it has also become important to discuss its legal and economic connotations. 

\subsection{Authorship, Copyright and Ethical Issues}

The case of Christie’s Belamy auction revealed many issues regarding the questions of authorship and copyright, as well as raised general discussions on the ethical considerations that have to be taken into account during production, promotion and sale of an AI artwork. In the case of the aforementioned auction, the artwork was presented as being autonomously produced by an AI system, yet the authors that created that system, nor the author of the code that was used to run the network, did not receive any formal acknowledgement. When an AI artworks gets sold for such an unexpectedly large price, who holds the right to profit from the sale becomes a very relevant question and triggers many discussions. McCormack et al. \cite{mccormack2019autonomy} provide a detailed overview of the problematic aspects of the ``Portrait of Edmond Belamy'' regarding authorship, authenticity and other important aspects of AI Art. Epstein et al. \cite{epstein2020gets} use the ``Portrait of Edmond Belamy'' case to explore how anthropomorphization of an AI system influences the perception of humans involved in the creation process. Stephensen \cite{stephensen2019towards} discusses the implication that the Belamy case has on the philosophical understanding of creativity. Colton et al. \cite{colton2018issues} discuss how human understanding of different notions of authenticity can be used to address computational authenticity. 

Despite the ongoing debates, most of the recent examples of sold AI artworks indicate that currently the authorship rights are attributed to the artist who produced the artwork using AI techniques, regardless of the narrative surrounding the creation process, e.g. the fact that the artwork was labelled as being made by an AI. While the idea that developers of a computational model are entitled to authorship rights can perhaps be dismissed in the same way in which it would be absurd to give credits for artistic photographs to the inventor of the camera, the question of data sources is a more complex one. Because part of the training data for AI Art generation using GANs could include copyrighted images, the final output would in that case involve someone else’s artistic contributions. This could of course be hardly noticeable in the final work, but still require an acknowledgement from an ethical perspective. To have a complete insight in all the possible contributions would require the disclosure of all the phases in the creative process. This is particularly relevant considering the current rising interest in purchasing AI Art which triggers possibilities of novel types of forgeries, e.g presenting images produced manually with image editing softwares as AI artworks. However, exposing all steps of their creative process is not something artists are really keen to do, precisely because specific procedural choices constitute the basis for originality and uniqueness of an AI artwork. In a comprehensive analysis of the issue of copyrights of artworks produced by creative robots, Yanisky-Ravid and Velez-Hernandez \cite{yanisky2018copyrightability} argue that confronting the challenges of the autonomous and automated content production calls for a reassessment of the meaning of originality. Several other recent articles indicate that copyright infringement in AI Artworks is becoming a relevant topic that needs to be systematically addressed \cite{gillotte2019copyright,eshraghian2020human, guadamuz2017androids}. 

Another important aspect of AI Art, as a growing subfield of digital art, is that it is bringing relevant changes to the contemporary art market. There is a rising trend to shift traditional art enterprises to corresponding online versions by establishing online galleries and online auctions. Furthermore, the very nature of digital artworks requires different approaches to ownership transactions than in the case of traditional physical artworks. In the last few years a new artistic movement emerged called CryptoArt which led to a great expansion of the so-called crypto art market which is based on the use of blockchain technology. Artworks are cryptographically registered with a token on a blockchain which allows them to be safely traded from one collector to another using crypto-currencies. Franceschet et al. \cite{franceschet2020crypto} present a detailed discussion on CryptoArt that includes viewpoints from artists, collectors, galleries, art scholars and data scientists involved in the system. Sidorova \cite{sidorova2019cyber} addresses the issue of digitalization of the contemporary art market and analyzes how cryptocurrency, blockchain, and artificial intelligence have the potential to contribute to the further development of online art trade.

\subsection{Perception of AI Art}

One of the major arguments for labelling generative AI systems as creative was the fact that the work they produced was indistinguishable from human-made art and perceived as surprising, interesting or aesthetically pleasing by a larger number of people. For example, the authors of the AICAN system performed a sort of visual Turing test to explore if people can tell the difference between AICAN- and human-made art \cite{elgammal2017can}. They conducted an experiment in which the mixed AICAN works with contemporary artworks shown at the Art Basel 2016 art fair. The outcome showed that 75\% of the time, people involved in the study thought that AICAN generated images were produced by human artists. However, Hong and Curran \cite{hong2019artificial} argue that the study conducted by Elgammal and colleagues used fewer than 20 participants and asked the participants directly if a work was created by humans or machines which might have introduced bias. Hong and Curran conduct their own experiment to study the perception of AI artworks which involved 288 participants. The results from their survey experiment show clear differences in evaluation between human-created artworks and AI-created artworks, with human-created artworks being rated significantly higher in properties such as ``composition,'' ``degree of expression,'' and ``aesthetic value.'' The authors conclude that the results of their study indicate that AI Art has yet to pass the Turing test for art. However, even if AI systems can, or will in the future, produce convincing artworks that resemble human-made art, that does not necessarily imply that the system itself should be perceived as truly autonomous or creative. In a detailed discussion on machine produced art, Ch’ng \cite{ch2019art} reflects on his own experience of an exhibition of artificially generated artworks by expressing his fascination with the ``illusion of consciousness depicted in these artworks''. Regardless of the current differences in framing the underlying process as advanced imitation or autonomous creativity, studying attitudes towards AI Art is an important novel research direction that can have strong implications on our general understanding of creativity. For example, an recent study by Wu et al. \cite{wu2020investigating} examined the explicit and implicit perceptions of AI-generated poems and paintings in the U.S. and China. The results of this study suggest that participants from the U.S. were more critical of the AI- than the human-generated content, both explicitly and implicitly. Chinese subjects were generally more positive about the AI-generated content, although they also appreciated human-authored content more than AI-generated. Previous studies also confirmed a negative bias in perception of AI-generated content in relation to human-made content \cite{hong2018bias, ragot2020ai}. Besides exploring human perception of AI Art, another topic of interest are visual characteristics of AI Art in the context of art history. Hertzmann introduces the concepts of visual indeterminacy as a specific stylistic property of AI Art created using GANs \cite{hertzmann2020visual}. Srinivasan and Uchino \cite{srinivasan2020biases} explore the biases in the generative art AI pipeline from the perspective of art history and also discuss the socio-cultural impacts of these biases.

\section{Conclusion and Future Outlook}

Current trends indicate that AI technologies will become more relevant in the analysis and production of art. In the last several years many universities have established Digital humanities (DH) master's and PhD programs to educate new generations of researchers familiar with quantitative and AI-based methods and their application to humanities data. We can expect that this will intensify the methodological shift from traditional towards digital research practices in the humanities, as well as result in a growing number of innovative research projects that apply large scale quantitative methods to study art-related historical questions. From the perspective of computer vision, there are still many practical challenges that need to be solved in order to assist researchers working on cultural digital archives. In particular, those are problems related to annotation standards, advanced object detection and retrieval, cross-depiction, iconographic classification, multi-modal alignment and image understanding. The use of deep neural network models was previously conditioned on the availability of large-scale datasets. By utilizing the concept of transfer learning and label-scarce techniques such as few shot learning, deep neural network models can be applied on smaller-sized dataset and employed for different fine-grained tasks and various image collections. Those kinds of approaches will probably be exploited by many future domain-specific digital art history projects. Besides employing deep learning models for enhancing research practices in art history, it is noteworthy to recognize the potential that tasks and data sources from the art domain have on the development of new computer vision and deep learning techniques. Digitized art collections are data sources of images that usually include rich contextual information related to the historical and technical aspects of their formation, but also represent a source of perceptually intriguing visual information which merges interweaving concepts of content and style. Because they comprise different layers of information, art collections represent a useful data source for addressing various and complex tasks of computational image understanding. 

In the context of art creation and production, AI technologies are starting to have an ever more important role. Not only in terms of digitally and AI produced art, but also in all the aspects of curation, exhibition and sale of traditional art as well. Having in mind the rapid shift of attention towards online platforms and digital showrooms due to the current global pandemic, the ongoing circumstances contributed to the already rising interest in crypto art and blockchain technologies which have the potential to significantly impact and transform the art market. Regarding the creation of art using AI technologies, in the last few years GAN-based approaches were dominating the AI Art scene. Recently, significant breakthroughs have been achieved in the development of multimodal generative models, e.g. models that can generate images from text. Technological advancement in this direction will probably have significant influence on the production and creation of art. Models that can translate data from different modalities into a joint semantic space represent an interesting tool for artistic exploration because the concept of multimodality is integral to many art forms and has always played an important role in the creative process. Furthermore, it is evident that the increasing use of AI technologies in the creation of art will have significant implications regarding the questions related to authorship, as well as on our human perception of art. With the development of AI models that can generate content which very convincingly imitates human textual, visual or musical creations, many of our traditional, as well as contemporary, theoretical and practical understandings of art might become challenged.  

\bibliographystyle{acm}  
\bibliography{aiart}
\end{document}